# Detecting Car Speed using Object Detection and Depth Estimation: A Deep Learning Framework

Subhasis Dasgupta
subhasis@praxis.ac.in
Praxis Business School, Kolkata

Arshi Naaz
arshinaaz1005@gmail.com
Praxis Tech School, Kolkata

Jayeeta Choudhury
jayeeta987@gmail.com
Praxis Tech School, Kolkata

Nancy Lahiri
lahiri.nancy1@gmail.com
Praxis Tech School, Kolkata

**Abstract-** Road accidents are quite common in almost every part of the world, and, in majority, fatal accidents are attributed to over speeding of vehicles. The tendency to over speeding is usually tried to be controlled using check points at various parts of the road but not all traffic police have the device to check speed with existing speed estimating devices such as LIDAR based, or Radar based guns. The current project tries to address the issue of vehicle speed estimation with handheld devices such as mobile phones or wearable cameras with network connection to estimate the speed using deep learning frameworks.

## I. Introduction

In today's transportation landscape, the ongoing issue of upholding road safety continues to be a major priority [1]. The ability to accurately measure and monitor the speed of vehicles on roads and highways is essential for ensuring compliance with speed limits, preventing accidents, and optimizing traffic flow. Studies have shown that speeding is a leading cause of traffic fatalities and injuries, emphasizing the importance of effective speed enforcement measures. By deterring speeding behaviour and encouraging compliance with speed limits, speed detection technologies contribute to the prevention of accidents and the mitigation of their consequences. This is where an effective speed monitoring system plays an important role.

Modern vehicle speed detection systems rely on a variety of technologies, including radar, LIDAR, and video-based systems. Radar-based systems use radio waves to measure the speed of vehicles by detecting the Doppler shift in the frequency of reflected signals. Lidar systems employ laser beams to calculate vehicle speed based on the time it takes for light pulses to reflect off a vehicle and return to the sensor. Video-based systems utilize cameras and image processing algorithms to track vehicle movement and estimate speed.

The current project falls into the third category, that is, video-based systems. To effectively address the challenge of speed detection, the proposed project integrates advanced technologies such as Convolutional Neural Networks (CNN), YOLOv8 [1], [2], [3], and OpenCV. The project effectively utilizes OpenCV to break down videos into individual frames, laying the groundwork for further analysis. YOLOv8, known for its object detection capabilities, is then employed to accurately identify vehicles in the video footage, thus improving the precision of speed detection. The study also involves utilization of depth estimation [4] to estimate the car speed with higher precision.

## II. Related Work

Vehicle speed detection is not a new concept. There are several methods available for estimating speed of a moving vehicle using LIDAR gun [5], Radar gun [6] or manual count method. But these methods are either costlier or region specific, i.e., speed can be estimated only in some specific points. That is why video based speed estimation methods are proposed because they are usually cost effective and can be quite accurate if trained with significantly large number of training data points. Most of the earlier methods used vehicle tracking [7] and estimating time requirements for a vehicle to cross a specific portion of the road [8]. All these methods relied heavily on the concepts of object detection using various pre-trained models such as YOLOv5, YOLOv8, SSD, Faster RCNN etc. However, these methods required a camera to be installed at a fixed position and at a certain height, leading to the same location specific constraint. That is why a need was felt to estimate vehicle speeds by any traffic police using his/her handheld mobile phones or any camera having an active internet connection. Hence, along with object detection, depth estimation was also thought be important in such situations. In this context, various works were studied and finally

MiDAS [9] python package was considered for depth estimation as this package is being maintained by the contributors quite actively.

**Data Collection**

Quality of data plays a very important role in the area of research. For the current study, the data was collected using primary method of data collection. For this study, around 100 data points were collected by carefully curating the videos of moving vehicles. The videos were captured by the hand held mobile phones having 640x480 resolution. For the present study, videos were collected in such a way so that only one car was there in the frame. Each video was either, shot from the front or from the side perspective. The videos varied in duration from 3 seconds to 7 seconds. During the data collection process, the vehicles' speeds were also recorded along with the video of the moving vehicle.

### III. Methodology

For the present study, data were collected in such as way so that a single car was present in a particular video. This project was not thought to be generalized for multi-vehicle scenario as vehicle tracking would have become important in multi-vehicle scenario. Once the curated videos were collected, the videos were sent through the OpenCV framework with YOLOv8 object detection model integrated. The purpose of the YOLOv8 was to create a tighter bounding box around the vehicle. However, during the process, it was seen that even the rear-view mirrors were getting identified as objects in some videos. Hence, a threshold confidence was needed to retain only the main vehicle. For this purpose, manual checks were performed to decide the threshold confidence probability supplied by the YOLOv8 model. The best threshold confidence was found to be 0.7, i.e., only those objects were to be retained for which the threshold confidence was above 0.7. The process of speed estimation of the car rested on the concept that as the car approaches the camera, the size of the bounding box should increase with time till it reaches some maximum value and, after which, as the car starts moving away from the camera, the size of the bounding box should decrease. If a car moves faster, within a small duration of time, the change in bounding box area would be higher than a slow moving car. Not only that, as the car approaches the camera with certain speed, the average distance of the pixels of the car from the camera should also decrease with time. This decrease should be dependent on the speed of the car. Hence, if 't' represents duration in seconds, $\Delta A$ represents the change in area of the bounding box and $\Delta D$ represents the change in average distance of the car from the camera, then speed can be modelled as a linear model as mentioned below in Equation 1:

$$Speed = \beta_0 + \beta_1 t + \beta_2 \Delta A + \beta_3 \Delta D + \epsilon \qquad (1)$$

Here, $\epsilon$ is the error. Calculation of $\Delta A$ was quite easy as the videos were clipped for different number of seconds and the frames per second was known. Hence, $\Delta A$ was found out by taking the difference between the bounding boxes of the car in the first frame of the video and the last frame of the video. However, finding out $\Delta D$ was a bit challenging. Depth estimation gives the distance of each pixel from the camera. The pixels nearer to the camera are red shifted and the pixels away from the camera are blue shifted. Within bounding box enclosed images, the background images were interfering with the distance estimation of the vehicle from the camera. To solve this problem to a greater extent, instead of bounding boxes, masks of the vehicle were considered with MiDAS output. To acquire the vehicle mask, Mask-RCNN was used. Mask RCNN works in the similar way as YOLOv8 to detect the object within the image but instead of creating a bounding box, this model creates a mask of the object to identify the object. The mask tries to create a closed contour of the object which is beneficial in different object detection scenarios. Thus, for initial and final average distance estimation of the vehicle, the first and the last frames of the videos were considered with both bounding boxes and vehicle masks. The entire process is depicted in Figure 1. To do a comparative analysis of object detection models, three models were used for object detection, i.e., YOLOv8, YOLOv5 and SSD. For comparison of models' performances $R^2$, $Adj\ R^2$ and $RMSE$ were considered. Both linear and polynomial regression were considered in this study to see the predictive power of the suggested models. To limit the number of features in polynomial regression, the maximum order was kept at 3.

### IV. Results and Analysis

As mentioned earlier, for this study 90 data points were collected. As the number of data points were less, sophisticated machine learning models were not utilized as those models would have required more data points to determine the pattern without getting overfit. The YOLOv8 model was found to be quite efficient in returning a tighter bounding box around the vehicles and hence calculating the area of the bounding was and easy task as YOLOv8 returned the coordinates of the bounding box also. Figure 2 and Figure 3 shows the outcome of

YOLOv8 of two such cars. For the present study, videos of cars were collected from two different perspectives, i.e., front and side.

It can be noted that in this study, due to lack of sufficient data, only pre-trained models were used. From the depth map it can be seen that the objects of interests were visible quite clearly and it was necessary to focus on the object of interests only and

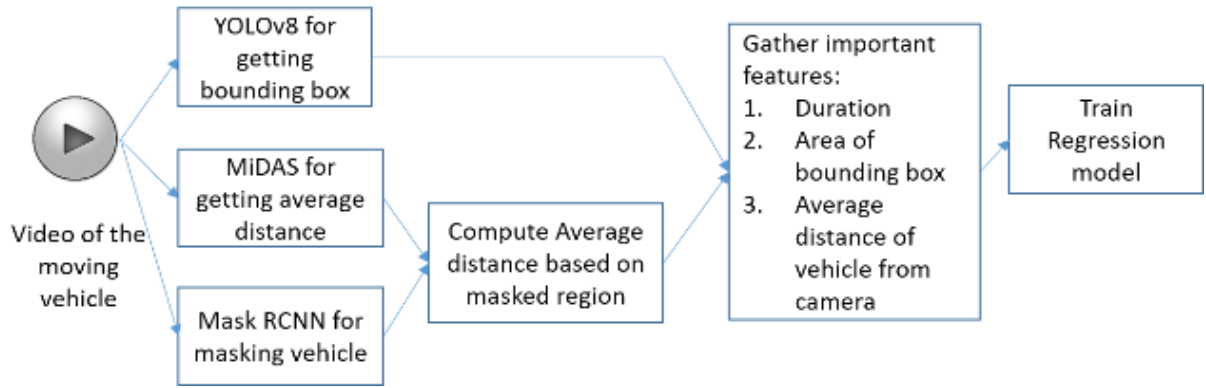

**Figure 1**. Flowchart of the model training process

The next part was associated with depth estimation. For this MiDAS package was used and the distances of each pixel from the camera was estimated. The distance map of the two cars are shown in Figure 4 and Figure 5. It is to be understood that the red colour portions represents the pixels nearer to the camera and the blue coloured portions represents the pixels more away from the camera.

to avoid any background image as much as possible. With bounding box, this was not possible as the box does not follow the contour of the object. Hence, it was decided to use object masking so that average distance of the object of interest only can be calculated more accurately.

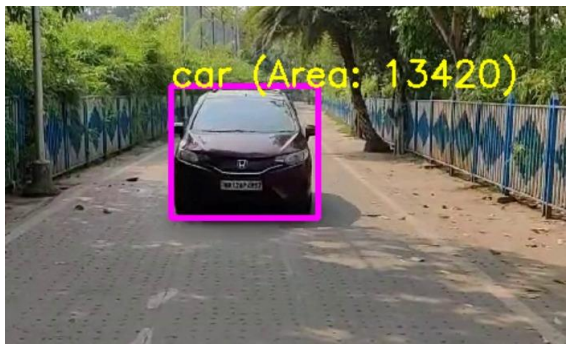

**Figure 2**. Object detection with bounding box and associated area (front view)

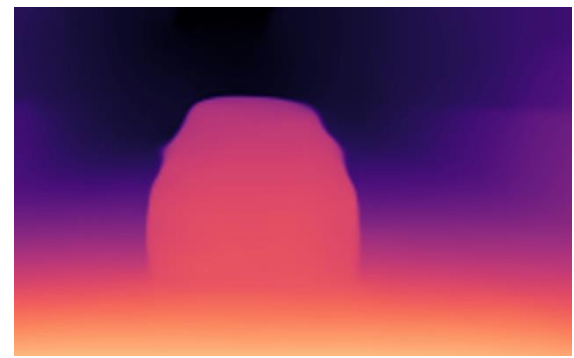

**Figure 4**. Depth map from the output of MiDAS for the front facing car

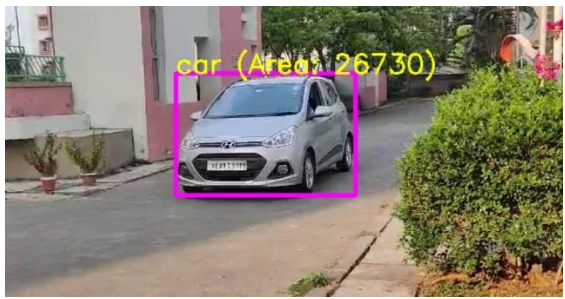

**Figure 3**. Object detection with bounding box and associated area (side view)

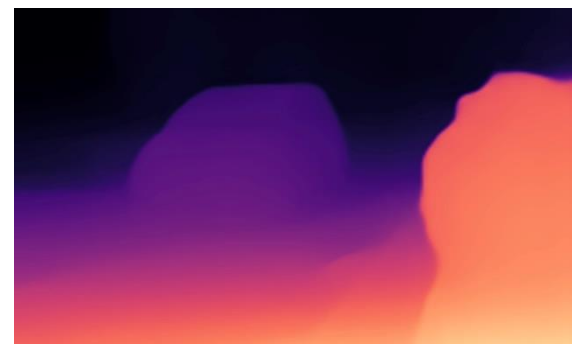

**Figure 5**. Depth map from the output of MiDAS for the side facing car

Thay is why pre-trained Mask RCNN model was used to mask the vehicle for more accurate distance estimation.

The output of the distance estimation of the vehicle using both MiDAS and Mask RCNN are shown in Figure 6 and Figure 7. In these figures, the masks try to match the contour of the object (the vehicle) as accurately as possible.

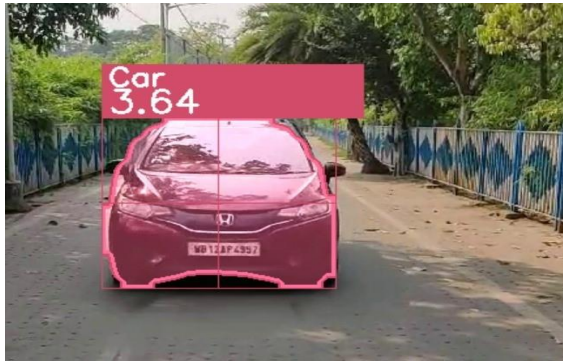

**Figure 6**. Average distance estimation using both MiDAS and Mask RCNN (front view)

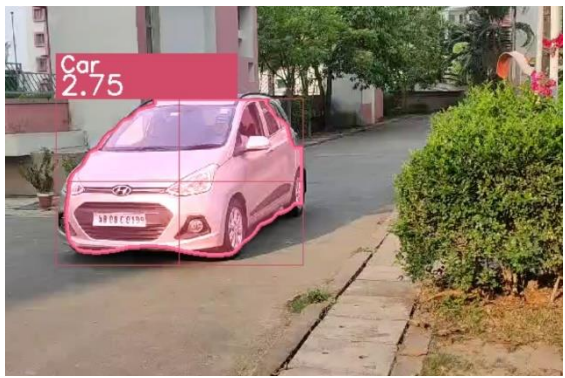

**Figure 7**. Average distance estimation using both MiDAS and Mask RCNN (side view).

After extracting the bounding box areas and the average distances of the vehicle in every frame, first and the last frame of the video were considered and the data points were created for building the regression model. The durations of the videos were varying from 2-6 seconds and for each video, the speed of the vehicle was known and was kept constant.

Initially, based on the extracted data, simple linear regression model was built. Two models were built, one without the information of change of distance based on MiDAS output and another with that extra information. However, the model's performance was substandard. The $R^2$ value observed was around 0.52 with MiDAS input and 0.4 without that input. This suggested that MiDAS input was quite important for speed estimation. The result also suggested that the relationship between the actual speed and the extracted features were not linear. Moreover, due to lack of data points, machine learning based models such as SVM or Artificial Neural Network was out of the scope. Hence, non-linear regression was tried with polynomial terms. In fact, polynomial feature extraction was done using python's scikit-learn package to create new features out of the existing features. Since polynomial regression can also get overfit, the dataset was split into 80-20 proportion for training and testing. With polynomial regression, a significant jump in $R^2$ value was observed and the Adj $R^2$ value was also quite high. The $R^2$ value jumped up to 0.81 with MiDAS input. Similar experiments were done with YOLOv5 and SSD for object detection. The comparative analysis of the experiments is shown in Table 1.

**Table 1**. Comparative analysis of performances of models

| Model (Object Detection) | Regression | Adj $R^2$ | $R^2$ | RMSE |
|---|---|---|---|---|
| YOLOv8 | Linear | 0.50 | 0.52 | 4.90 |
|  | Non-Linear | **0.74** | **0.81** | **2.24** |
| YOLOv5 | Linear | 0.45 | 0.47 | 5.18 |
|  | Non-Linear | 0.66 | 0.76 | 2.50 |
| SSD | Linear | 0.46 | 0.48 | 5.28 |
|  | Non-Linear | 0.60 | 0.72 | 2.71 |

It can be seen clearly from the above table that the YOLOv8 models for object detection produced the best result with the lowest RMSE score. Feature importance of the best model is shown in Figure 8. As per the figure, the most important feature is difference between the average distance of the car from the camera in the initial and in the final frame. "Diff^1" denotes this feature. This is followed by the square of the difference in the bounding boxes in the initial and the final frame of the video. "Area Diff^2" denotes this feature. Likewise, other features are mentioned on the image of feature importance.

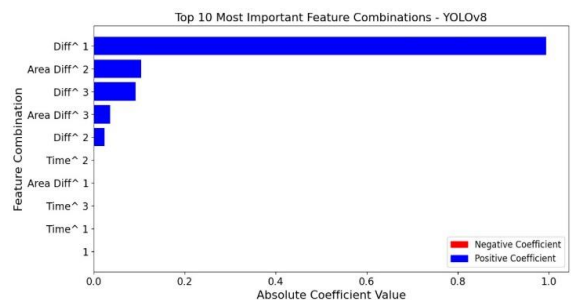

**Figure 8**. Feature importance plot of the best model

The relationship between actual speed and predicted speeds are shown in Figure 9 and Figure 10. The plot for non-linear model also shows the performance of the model on the test dataset with red dots. The errors in predictions are much lesser in this case also. The final predictions are shown in Figure 11 as a sample output. The values appearing at the right hand lower corner of an individual image is the frame number and the values appearing at the top are the actual vs the predicted speed.

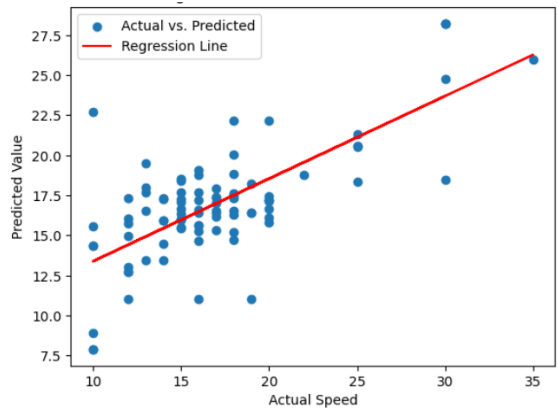

Figure 9. Actual vs Predicted speed for OLS model

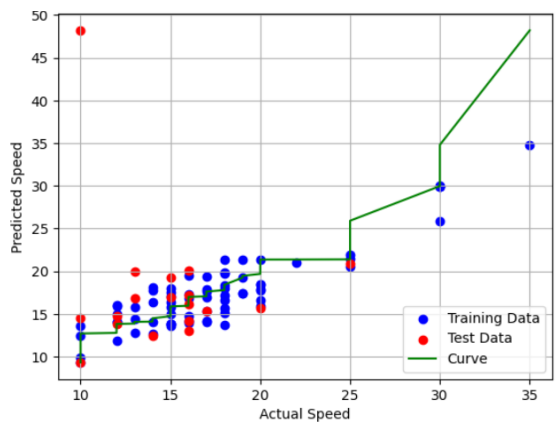

Figure 10. Actual vs Predicted speed for the non-linear model

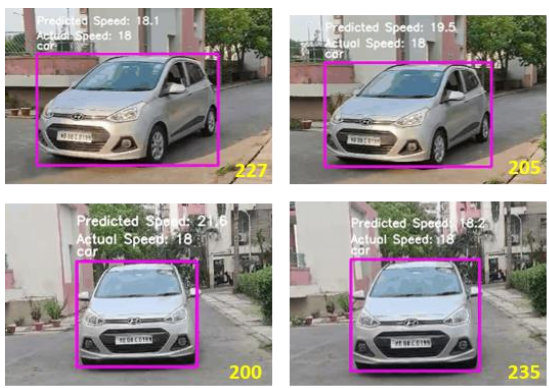

Figure 11. Sample model output of a moving car in a two videos at different frames.

It can be seen from the image that as the car is away from the camera, the speed estimation is less accurate. But as the vehicle starts closing on the camera, the accuracy of the model increases to a very good extent. For example, as per the above image, for the front viewing car, when the car was somwhat away (i.e., frame 200), the actual speed was 18 kmph whereas the predicted speed was 21.6 kmph. But, when the same car reached closer to the camera (i.e., frame 235), the predicted speed became 18.2 kmph. The similar behaviour is also observed while predicting car speed with side perspective. Hence, for actual use case, speed of the vehicle can be finalized by taking into consideration the frames where the average distance of the vehicle from the camera is minimum or closer to minimum.

## V. Conclusions

The present work focused on estimating vehicle speed using pre-trained deep learning models. It was found out that if depth estimation is done properly, the accuracy of prediction becomes more accurate. Moreover, accurate bounding box prediction leads to better speed estimation and YOLOv8 model could produce more accurate bounding boxes around the vehicles and that lead to much better predictive power of the non-linear regression model. However, it is also to be noted that the model may not be a good choice if speed is to be estimated at every frame of the video because even though YOLOv8 is quite fast in doing analysis, Mask RCNN would create the bottleneck due to its slower speed compared to YOLOv8. Hence, this process may be adopted if speed analysis is done every second. The other bottleneck could be the case when the vehicle is moving very fast (say above 100 kmph). In that case the shutter speed of the camera will play a pivotal role to get sharp images of the vehicle even at high speed. Finally, the project did not involve capturing the registration number of the vehicle and hence, in future work, this project can be integrated with vehicle registration plate identification with deep learning for better law enforcements.

### Acknowledgement

The researchers acknowledge the contributions made by Bidisha Sadhu, Siddharth Tiwari and Matta Sai Mohan Ranga in preparing the data as per the instructions provided by the researchers.